\documentclass[11pt,a4paper]{article}

\usepackage{booktabs} 
\usepackage{emnlp2017}
\usepackage{times}

\usepackage{latexsym}
\usepackage{courier}
\usepackage{dsfont}

\usepackage{balance}
\usepackage{latexsym}
\usepackage{times}
\usepackage{changepage}
\usepackage{latexsym}
\usepackage{algorithm2e}
\usepackage{amsmath}
\usepackage{epsfig}
\usepackage{epstopdf}
\usepackage{boxedminipage}
\usepackage{algpseudocode}
\usepackage{graphicx}
\usepackage{tabularx}
\usepackage{times}
\usepackage{multirow}
\usepackage{latexsym}
\usepackage{tablefootnote}
\usepackage{bbm}
\usepackage{comment}
\usepackage{svg}
\usepackage{listings}
\usepackage{setspace} 

\emnlpfinalcopy

\definecolor{armygreen}{rgb}{0.29, 0.33, 0.13}
\definecolor{cadmiumgreen}{rgb}{0.0, 0.42, 0.24}

\begin{document}

\title{Towards Understanding and Answering Multi-Sentence Recommendation Questions on Tourism}

 \author{Danish Contractor\thanks{This work was carried out as part of  PhD research at IIT Delhi. The author is also a regular employee at IBM Research. } \\IIT Delhi \& IBM Research AI\\ New Delhi, India\\     {\tt dcontrac@in.ibm.com} \And Barun Patra \\ IIT Delhi \\New Delhi, India \\ {\tt cs1130773@cse.iitd.ac.in}
        \AND Mausam \and Parag Singla \\ IIT Delhi \\ New Delhi, India \\ {\tt \{mausam,parags\}  @cse.iitd.ac.in}}

\maketitle
\begin{abstract}

We introduce the first system towards the novel task of answering complex multi-sentence recommendation questions in the tourism domain. 
Our solution uses a pipeline of two modules: question understanding and answering. 
{\color{black} For question understanding, we define an SQL-like query language that captures the semantic intent of a question; it supports operators like subset, negation, preference and similarity, which are often found in recommendation questions.} {\color{black}We train and compare traditional CRFs as well as bidirectional LSTM-based models {\color{black}for converting a question to its semantic representation.} We extend these models to a semi-supervised setting with partially labeled sequences gathered through crowdsourcing. We find that our best model performs semi-supervised training of BiDiLSTM+CRF with hand-designed features and CCM\cite{CoDL} constraints.
}
{\color{black}Finally, in an end to end QA system, our answering component converts our question representation into queries fired on underlying knowledge sources. Our experiments on two different answer corpora demonstrate that our system can significantly outperform baselines with up to 20 pt higher accuracy and 17 pt higher recall.}

\end{abstract}

\section{Introduction}

\begin{figure*}[ht]
 {\footnotesize
 \center
   \includegraphics[scale=0.45]{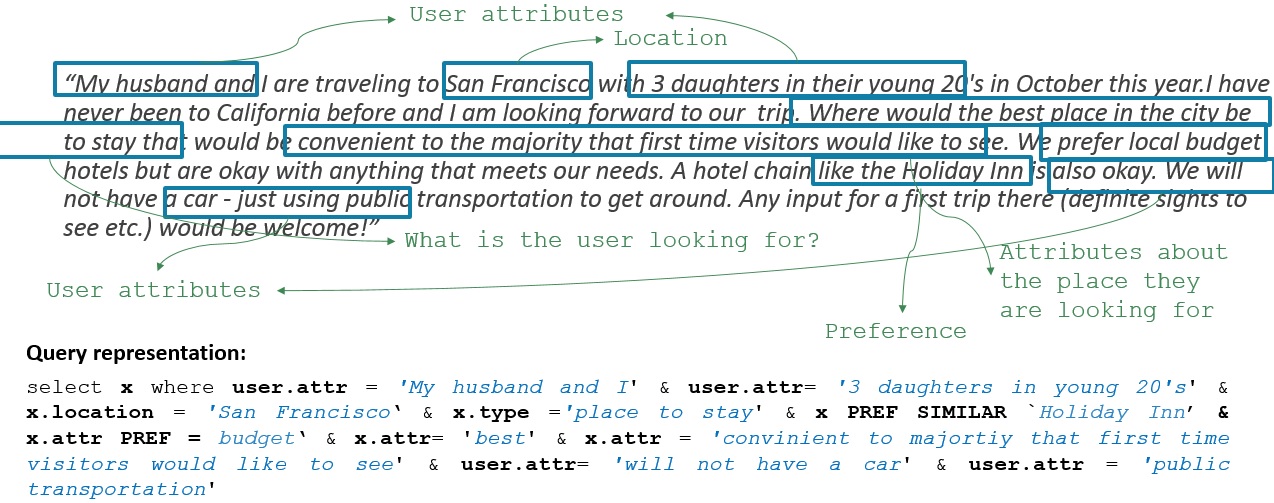}
   \vspace*{-2ex}
   \caption{An entity-seeking MSRQ and its corresponding RQL representation.}\label{fig:queryExample}
 }
 \end{figure*}

We are motivated by the goal of building an information agent for tourists -- one that would perform various roles of a travel agent, such as helping decide the city to visit, recommending points of interest, finding travel routes, and even creating optimized itineraries. Our paper develops a key component of such an agent -- a QA system for directly answering {\em recommendation questions}. As a first step, we focus our paper on questions that are {\em entity-seeking}, i.e., expect one or more entities as answer. These include the large fraction of tourist questions that ask for hotels, restaurants, points of interest and other services that would serve a user's specific needs the best. Figure \ref{fig:queryExample} shows an example of such a question, where the user is interested in finding a hotel that satisfies some constraints and preferences;  an {\em answer} to this question is thus the name of a hotel (entity). 

A preliminary analysis of such questions from popular tourism forums reveals that almost all of them contain multiple sentences -- they often elaborate on a user's specific situation before asking their question. We name these MSRQs -- {\em multi-sentence recommendation questions}. An answering system needs to retrieve answer entities from background knowledge sources that may have information about each candidate entity. This includes review sites like TripAdvisor, Booking.com, travel guides such as WikiTravel,\footnote{\em http://www.wikitravel.org} or  online services like Google Places.\footnote{\em https://developers.google.com/places/} 

Understanding MSRQs raises several novel challenges. MSRQs use informal language, express a wide variety of intents and requirements in each question, and express user preferences and constraints in addition to those for the answer. The questions can be unnecessarily belabored requiring the system to reason about what is important and what is not. Moreover, the querying module needs to incorporate the various constructs found in recommendation questions.

\subsection{Contributions}

We present the first system for the novel task of answering {\color{black} entity-seeking} MSRQs from background corpora in the tourism domain. We make three main technical contributions: a query language to represent MSRQs, a question understanding module to parse the question into our language, and answering systems that perform retrieval over knowledge sources to return answer entities.

\noindent
{\bf Query Language: } {\color{black} For question understanding, we define a recommendation question language (RQL) that captures the semantic intent of an MSRQ. RQL is an SQL-like language with operators chosen to cover various entity-seeking MSRQs. It expresses both attributes of the answer and the user. We construct a dataset of MSRQs and their RQL representations (9200 annotated tokens), used as training data.} 

\noindent {\bf Question Understanding: }
We take a sequence labeling approach for question understanding. The current state of the art for sequence labeling uses bidirectional LSTMs with a final CRF layer \cite{LSTMCRF}. However, because our dataset is relatively small, it was not clear a priori whether a purely neural solution for sequence labeling will be competitive with the more traditional feature engineering methods. In response, we perform extensive experiments and combine various approaches to construct our best model. It includes (1) neural features, (2) hand-designed features (3) constraints capturing additional domain knowledge, and (4) semi-supervised learning over crowdsourced data. Our final model runs the Constraint Driven Learning Algorithm (CoDL)\cite{CoDL} for semi-supervised learning over a BiDiLSTM+CRF with additional features and constraints modeled using constraint conditional modeling (CCM)\cite{CoDL}. We evaluate the incremental value of each addition to the model.

\noindent
{\bf End-to-End QA Experiments: }
After parsing the question into RQL, our corpus-specific query generators convert the parse into queries that can be fired on the knowledge source for generating answers. 
{\color{black} Since answering MSRQs is a novel task, there are no direct baselines}. We therefore, 
compare our system against a QA system - WebQA \cite{complexwebQA2016} which was designed to handle multi-sentence questions; the original WebQA system returned passages as answers, but we adapt it to return entity answers from our knowledge sources for fair comparison. We use two knowledge sources: (1) an offline corpus of reviews for about half a million entities crawled from Google Places, TripAdvisor, WikiTravel and Booking.com; (2) full online Google Places API. 
We find that our RQL-based QA dramatically outperforms baselines obtaining 20 pt accuracy, and 17 pt recall improvements. 

\section{Related Work} \label{sec:related}

To the best of our knowledge, we are the first to explicitly {\em understand} and {\em directly answer} multi-sentence recommendation by returning entities using a background corpus. 
Our work is related to the research in question understanding and other forms of QA.

\noindent{\bf Question Answering Systems: } There are two common approaches for QA systems -- joint and pipelined, {\color{black} both with different advantages}. The joint systems usually train an end-to-end neural architecture, with a softmax over candidate answers (or spans over a given passage) as the final layer \cite{QuizBowl,squad}. {\color{black}Such systems can be rapidly retrained for different domains, as they use minimal hand-constructed or domain-specific features. But, they require huge amounts of labeled QA pairs for training.} 

In contrast, a pipelined approach \cite{OQA, PARASEMPRE,Paralex,Kwiatkowski,complexwebQA2016,LiveQAWangN16} divides the task into two components -- question processing (understanding) and querying the knowledge source. 
Since each of these are simpler sub-problems, such methods can be built with relatively less training data, but require more annotation efforts per domain.

It is important to note that for answering an MSRQ, the answer space can include thousands of candidate entities per question, with large unstructured review documents about each entity that help determine the best answer entity. We briefly summarize popular approaches in QA systems for easy comparison of our work with existing literature in Table \ref{tab:related_QA}: QA systems and can be broadly classified based on (a) type of questions they answer (b) nature of KB/Corpus used for answering (c) nature of answers returned by the answering system

\begin{table*}[]
\hspace{-1cm}
{\small
\begin{tabular}{|c|p{5.5cm}|p{3cm}|p{5cm}|}
\hline
\textbf{Question Type}                    & \multicolumn{1}{|c|}{\textbf{Knowledge Type}}                               & \multicolumn{1}{|c|}{\textbf{Answer Type}} & \multicolumn{1}{|c|}{\textbf{Related Work}} \\ \hline \hline
          & Structured (eg. DBPedia, Freebase)                                        & Entity                                   &  \cite{WWW2017Factoid,bordes2,bordes2015large}                                         \\ \cline{2-4}
 \multirow{6}{*}{Single Sentence}                                         & Structured (Open IE style KBs)                                           & Entity                                   &      \cite{OQA,PARASEMPRE}                                     \\ \cline{2-4}
 &Structured + Unstructured (Open IE style KBs with supporting text passages on entities ) & Entity & \cite{OpenIEWithClubWebQA} \\ 
 \cline{2-4}
                                          & Structured (Databases)                                                     & Tables/ Table rows                       &  \cite{Athena,nlidb-researchsurvey}                                      \\ \cline{2-4}
                                          & Unstructured                  & Text Spans                               & \cite{squad,NEWSQA,trivedi-2017-lcquad-iswc,WikiPediaReader,TriviaQA}                                           \\ \cline{2-4}
                                          & Unstructured                  & Text Passages                            & \cite{complexwebQA2016,LiveQAWangN16,LiveQAWangN15}                                           \\ \cline{2-4}
                                          & Multiple choice answers                                                   & Answers from specified choices           &   \cite{MCQA,ComplexQAIE}                                        \\ \hline \hline
\multirow{3}{*}{Multi-sentence} & Unstructured              & Text (Answer) passages                   &    \cite{unanswered-cqa,romeo2016neural,cqa-survey,reanking}\\ \cline{2-4}
                                          & Unstructured (QA pairs)                                            & Entity                                   &                          \cite{QuizBowl}                 \\ \cline{2-4}
                                          & {\bf Semi-structured meta-data + Unstructured (Entity Reviews)} & {\bf Entity}                                  & \textbf{Our work}   \\ \hline                      

\end{tabular}
}
\caption{Related work: QA}
\label{tab:related_QA}

\end{table*}

The problem of directly returning answers to questions from background knowledge sources has been studied, but primarily for single sentence factoid-like questions \cite{OQA,PARASEMPRE,TAQA,QuASE,Athena,ComplexQAIE,WWW2017Factoid}. Reading comprehension tasks  \cite{squad,NEWSQA,TriviaQA,trivedi-2017-lcquad-iswc} require answers to be generated from  unstructured text also only return answers for simple single-sentence questions. Other works have considered multi-sentence questions, but in different settings, such as the specialized setting of answering multiple-choice SAT and science questions \cite{ai2, aristo,ComplexQAIE,MCQA}, mathematical word problems \cite{mathSolver}, and textbook questions \cite{mrinmaya}. Community QA systems \cite{reanking,shen2015word,convForCQA,lstmForCQA} match questions with {\em user}-provided answers, instead of entities from background knowledge-source. IR-based systems \cite{LiveQAWangN16} query the Web for open-domain questions, but return long (1000 character) passages as answers; they haven't been tested on recommendation questions. 
The techniques that can handle MSRQs \cite{complexwebQA2016,LiveQAWangN16} typically perform retrieval using keywords extracted from questions; these do not understand the questions well and can't answer many tourism questions, as our experiments show. The more traditional solutions (e.g., semantic parsing) that parse the questions deeply can process only {\em single}-sentence questions \cite{OQA,PARASEMPRE,Paralex,Kwiatkowski}. 

Finally, systems such as QANTA \cite{QuizBowl} also answer complex multi-sentence questions but their methods can only select answers from a small list of entities and also require large amounts of training data with redundancy of QA pairs. In contrast, the subset of Google Places we experiment with has close to half a million entities. Further, in our task, the reviews about each entity are significantly longer\footnote{Reviews for each entity are contactenated to serve as background information about that entity, resulting in documents ranging in length from a few hundred to a few thousand sentences.} than passages (or similar length articles) that have traditionally been used in QA tasks and it is only recently that the task of QA via neural machine comprehension of long documents has been proposed \cite{trivedi-2017-lcquad-iswc}.

\noindent{\bf Semantic Representation of Questions:} QA systems use a variety of different intermediate semantic representations. Most of them, including the rich body of work in NLIDB and semantic parsing, parse {\em single} sentence questions into a query based on the underlying ontology or DB schema \cite{nlidb-researchsurvey,Athena,Luke2,liangthesis,trivedi-2017-lcquad-iswc}. Open QA \cite{OQA} uses an open-domain representation for factoid single-sentence QA.

Recent works build neural models that represent a question as a continuous-valued vector \cite{bordes1,bordes2,reddyACL,WSDM2016,WSDB2016b}. Some systems rely on IR and do not construct explicit semantic representations at all \cite{QuASE,complexwebQA2016}; instead, they rely on selecting keywords from the question for querying. They can handle multi-sentence questions, but do not understand questions deeply. To the best of our knowledge no question parser has been developed for MSRQs.

\begin{figure}[ht]
{\footnotesize
\center
  \includegraphics[scale=0.58]{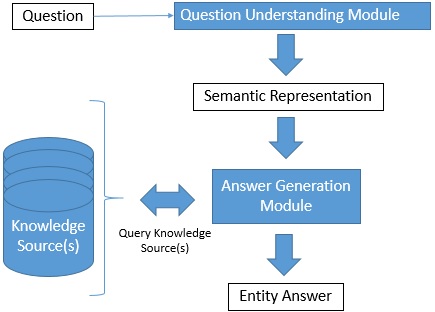}
  \caption{Schematic Representation of the system}\label{fig:qa-system}
}
\end{figure}

\section{System Architecture} \label{sec:sys}
Our QA system broadly consists of two modules (see Figure \ref{fig:qa-system}): question understanding, and answer generation. As motivated earlier, the modularized two-step architecture allows us to tackle different aspects of the problem independently. The semantic representation generated by the question understanding module is generic and not tied to a specific corpora or ontology. This allows the answering module to be optimized efficiently for each knowledge source, as well as allows the integration of multiple data sources, each with their own schema and strengths for answering. We first describe the details of our question representation (Section \ref{sec:rep}). Further sections describe the question understanding (Section \ref{sec:label}) and answer generation (Section \ref{sec:end-task}) modules in detail. {\color{black}Examples of some user questions and the answer entities returned by our system are shown in Table \ref{tab:questions}.}

\begin{center}
\begin{figure}
\vspace*{-3ex}
{\footnotesize
\begin{lstlisting}   
Q-> select x where C
C-> C | (C and C) | (C or C)
C-> L R T | L' near R T
R-> = | P = | in {T} | in [T] 
T-> T,T' | T'
T'-> `<phrase>'
P-> pref | similar | in | not |
P-> P P 
L-> x.attr | user.attr | x.type |
    x | x.location | user.location
L'-> x.location
\end{lstlisting}
\vspace*{-3ex}
\caption{\label{fig:CFG} CFG rules for our query representation}
}
\end{figure}
\vspace{-0.5cm}
\end{center}

\section{RQL Representation} \label{sec:rep}

{\color{black} Since we build the first system of its kind, we need to balance representation expressiveness and its answerability. For our first version, we make the assumption that the MSRQ  is asking only {\em one} final question, and that the expected answer is one or more entities. This precludes Boolean, comparison, `why'/`how', and multiple part questions. We now describe RQL, our language for representing such MSRQs.

We choose a relatively {\em open} question representation for RQL. It makes minimal assumptions about the answering knowledge sources and therefore, minimizes schema or ontology specific semantic vocabulary. 
Another advantange is that RQL with small changes could be rapidly adapted to a non-tourism domain (see Table \ref{tab:domainExamples} for examples of RQL queries for automobiles and electronics domains). At the same time, we note that RQL can easily be extended to include more schema-specific semantic labels, if required.}

\begin{table*}[ht]

\centering
\tiny
\begin{tabular}{|p{1cm}|p{4cm}|p{3cm}|}
\hline
\textbf{Domain}                                           & \multicolumn{1}{|c|}{\textbf{Question}}                                                                                                                                                                                                                                                                                & \multicolumn{1}{|c|}{\textbf{RQL Representation}}                                                                                                                                    \\ \hline \hline
\multicolumn{1}{|l|}{\multirow{2}{*}{\textbf{Automobiles}}} & I want to buy a sedan in diesel version and budget is USD 30,000-40,000. looking for one with basic luxury, nothing too fancy. Which one is best:?                                                                                                                                                                   & select x where x.type="sedan" \& x.attribute="diesel version" \& x.attribute="USD 30000-40000" \& x.attribute="basic luxury" \& x.attribute="nothing too fancy                     \\ \hline
\multicolumn{1}{|l|}{{\textbf{Automobiles}}}                                      & Can anyone suggest to me a reliable brand of a tyre pressure guage and pump? An estimate of their approx costs and place of availability in Delhi would be preferrable.                                                                                                                                              & select x where x.type="place of availability" \& x.attribute="tyre pressure guage and pump" \& x.attribute="reliable brand" \& x.location PREF ="Delhi                             \\ \hline
\multirow{2}{*}{\textbf{Electronics}}                     & My 15 year old Broksonic TV is dying, so I am needing to buy a replacement. Want an LCD TV about the same size (20"), with good picture clarity and sound quality. Must have composite (RCA) or component connectors to fit my DVD recorder, and VCR. Looking to buy from Amazon, TigerDirect, etc. Recommendations? & select x where x.type="LCD TV" \& x.attribute="20\", with good picture quality" \& x.attribute="composite (RCA) or component connectors" \& x.attribute="from Amazon, TigerDirect" \\ \hline
        {\textbf{Electronics}}                                                  & I've done a search and concluded that I can't afford the best washing machine (Miele etc) so how about some recommendations for a good quality front loader,7.5 kg and up to $1000$.Thanks.                                                                                                                          & select x where x.type="washing machine" \& x.attribute="good quality front loader" \& x.attribute="7.5 Kg" \& x.attribute="upto \$1000",x NOT SIMILAR "Miele"                     \\ \hline
\end{tabular}
\caption{RQL representations in different domains}
\label{tab:domainExamples}
\end{table*}

We illustrate RQL's representation choices by means of an example (Figure \ref{fig:queryExample}). Here, the user is interested in finding a hotel that satisfies some constraints and preferences. The question includes some information about the user herself, which may or may not be relevant for answering. 

RQL resembles an SQL-like language. Since each question has an entity (or more) as an answer, it denotes the desired answer by $x$. Each answer will have a type (referred to by the semantic label $x.type$), e.g., `place to stay'. Many tourism questions may be about facilities, which may have an $x.location$. 
To accommodate other characteristics of the answer entity, RQL defines  $x.attribute$ -- any phrase describing the answer that is not type or location is marked as an attribute.
Users often expect personalized answers and explain their individual situation in their question. To accommodate aspects of a user that may be important while answering, RQL defines a special entity  called $user$. It maintains $user.attribute$ and $user.location$ for user's features; `three daughters in their young 20s' will be marked $user.attribute$ in our example.

An analysis of tourism forum questions reveals that RQL can adequately represent almost all tourism questions that satisfy our assumptions. {\color{black}Notice the limited semantic vocabulary for candidate answers (type, location, attribute) -- this aligns with our goal of making minimal assumptions about the knowledge sources.}

\noindent {\bf Operators:}
Another key feature of RQL is that it maximizes the coverage of 
{\color{black} common operators} found in a recommendation question, so that a robust down-stream QA or IR system can meaningfully answer it.
In addition to standard logical connectives like  AND, OR and NOT (for example, the phrase \textit{``not very spicy''} may be represented as $x.attribute$ NOT =  `very spicy'), RQL also defines four more operators (PREF, NEAR, IN, and  SIMILAR) to represent common constructs in MSRQs.

PREF expresses a preference that is not a constraint, e.g., ``I would prefer to eat sushi'' ($x.attribute$ PREF =  `sushi'). The NEAR operator is used when a user requires recommendations that are geographically close to a location specified in the question (e.g ``\emph{near Salzburg}'' will be annotated as $x.location$ NEAR `Salzburg'). 

SIMILAR is used when a user mentions similar entities. For instance, {\it ``I have been to Red Hoods and wanna visit a similar place''} will be annotated as $x$ SIMILAR = `Red Hoods'. We note that mention of entities such as `Red Hoods' can be very informative, since these typically represent siblings of the answer -- instances of the type of the desired answer. We name these as {\em sibling entities}.

Finally, IN is used when a user explicitly provides a list of sibling entities among which she wants an answer. An example is {\it ``I have shortlisted Red Hoods, Cafe China and Royals. Help me''}; RQL construct will be  $x$ IN \{`Red Hoods', `Cafe China', `Royals'\}. Curly brackets denote an enumerated set. RQL also uses a special square bracket symbol to denote a range. For example, {\em ``location between New York and New Jersey"} will translate to $x.location$ IN [`New York', `New Jersey']. These operators may also be nested, for example, {\em``preferably not Red Hoods or Royals''} will be $x$ PREF NOT IN \{`Red Hoods', `Royals'\}.

Formally, RQL language is defined in Figure \ref{fig:CFG}. A query $Q$ is generated by a set of clauses $C$. Each clause $C$ can generate a label $L$ with operator $R$ and a terminal `$<$phrase$>$' generated by $T$ and $T'$. A separate rule with $L'$ is written for NEAR operators since they only support $x.location$. 

Two expert annotators with background in NLP annotate 150 user questions (9200 annotated tokens) with their RQL representations. These questions are chosen randomly from a popular tourism forum site. The annotators resolve their differences in person to produce a combined labeled set, which serves as training data for question understanding in the rest of the paper.

\section{Question Understanding}
\label{sec:label}

Before describing implementation details of our question understanding component, we present some background on Constrained Conditional Models (CCMs)\cite{CoDL} and BiDiLSTM+CRF\cite{LSTMCRF} as these are at the core of our question understanding component.

\subsection{Background on Sequence Labeling} \label{sec:background}

\noindent {\bf Constraint Conditional Models}
(CCMs) extend CRFs \cite{crf} by allowing an expert to express domain knowledge through hard or soft constraints. CCMs use an alternate learning objective expressed as the difference between the original CRF log-likelihood and a constraint violation penalty \cite{CoDL}:

{\small
\begin{equation*}
\sum_i w^{T}\phi({\bf x^{(i)}},{\bf y^{(i)}}) - \sum_i
\sum_k \rho_k d_{C_k}({\bf x^{(i)}},{\bf y^{(i)}})
\end{equation*}
}
Here, 
${\bf x^{(i)}}$ is the $i^{th}$ sequence and ${\bf y^{(i)}}$ its labeling. 
$\phi$ and $w$ are feature and weight vectors respectively. $d_{C_k}$ and $\rho_k$ denote the violation score and weight associated with $k^{th}$ constraint. 
The $w$ parameters are learned analogous to a vanilla CRF and computing $\rho$ parameters resorts to counting. Hard constraints have an infinite weight. Inference in CCMs is formulated as an Integer Linear Program (ILP); see Chang et al.\shortcite{CoDL} for details.


\noindent{\bf Constraints driven learning} (CODL) is a semi-supervised iterative weight update algorithm, where the weights at each step are computed using a combination of the models learned on the labeled and the unlabeled set \cite{CoDL}. CoDL's weight update equation is: 

{\small
\begin{equation*}
(w^{(t+1)},\rho^{(t+1)}) = \gamma(w^{(0)},\rho^{(0)}) + (1-\gamma) \rm{Learn}(U^{(t)})
\end{equation*} \label{updateEq}
}
Here, $t$ denotes the iteration number. $\rm{Learn}$ function learns the parameters of the model on examples supplied as its argument. The parameters at iteration $t=0$ are learned only using the labeled data. $U^{(t)}$ denotes the unlabeled set whose values have been filled in using parameters at iteration $t$.  $\gamma$ controls the relative importance of the labeled and unlabeled examples. 

\noindent{\bf Bidirectional LSTM based CRF} is a state of the art neural approach for sequence labeling \cite{LSTMCRF}. The output of BiDi LSTM network at each time step feeds into a CRF layer. The consecutive outputs of the LSTM states are connected to each other in the CRF layer and consist of a state transition matrix that contain the probabilities of transitions between output labels. 

It can be seen as a combination of neural feature engineering and CRF's joint inference. 
\subsection{Semantic Labeling of Questions} 
\label{sec:labeling}
{\color{black} We now discuss our approach to parse an MSRQ into its RQL representation. We could use a  full-blown CFG parsing approach. But given that the context-free component of our grammar is limited (most clauses are conjunctive), we, instead, use a combination of sequence labeling and operator post-processing. }
Our token-level label set directly corresponds to semantic labels in RQL:  \{{\em x.type, x.attribute, x.location, x.sibling, user.attribute, user.location, other}\}. 
Here, $x.sibling$ refers to sibling entities and  $other$ label captures all tokens not assigned any of the semantic labels. For now, we mark operator words as $other$ and handle them separately as a post-processing step using a set of lexical rules.  

{\color{black}Our sequence labeling task  dataset is relatively small. Because neural approaches are often more effective in large data settings, we experimented with both 
solutions -- traditional CRFs and BiDiLSTM CRF.
We further improve performance using CCM constraints and crowdsourcing more data.}

\subsubsection{Supervised Labeling}
\vspace{-1ex}

\noindent{\bf Conditional Random Field: } 
We first pose the sequence labeling task as a single linear chain CRF \cite{crf} over the MSRQ. We implement a number of features as follows. (a) Lexical features for capitalization, indicating numerals etc., token-level features based on POS and NER (b) hand-designed $x.type$ and $x.attribute$ specific features. These include indicators for guessing potential types, based on targets of WH ({\it what}, {\it where}, {\it which}) words and certain verb classes; dependency parse features that aid in attribute detection, e.g., for every noun and adjective, an attribute indicator feature is on if any of its ancestors is a potential $type$ as indicated by type feature; indicator features for descriptive phrases \cite{naaclpaper}, such as adjective-noun pairs.  (c) For each token, we include cluster ids generated from a clustering of word2vec vectors \cite{word2vec} run over a large tourism corpus. (d) We also use the counts of a token in the entire post, as a feature for that token \cite{complexwebQA2016}. 

\noindent{\bf Constrained Conditional Model: }
Since we label multi-sentence questions, we need to capture patterns spanning across sentences. One method of doing so would be to model these patterns as features defined over non-adjacent tokens (labels). But this can make the modeling quite complex. Instead, we model them as global constraints over the set of possible labels using CCMs. We design the following constraints: (i) type constraint (hard): every question must have at least one $x.type$ token, and (ii) attribute constraint (soft), which penalizes absence of an $x.attribute$ label in the sequence. (iii) a soft constraint that prefers all $x.type$ tokens occur in the same sentence. The last constraint helps reduce erroneous $x.type$ labels but allows the labeler to choose $x.type$-labeled tokens from multiple sentences only if it is very confident. Thus, while the first two constraints are directed towards improving recall, the last constraint helps improve precision of $x.type$ labels. 
\begin{figure}[ht]
{\footnotesize
\center
  \includegraphics[scale=0.45]{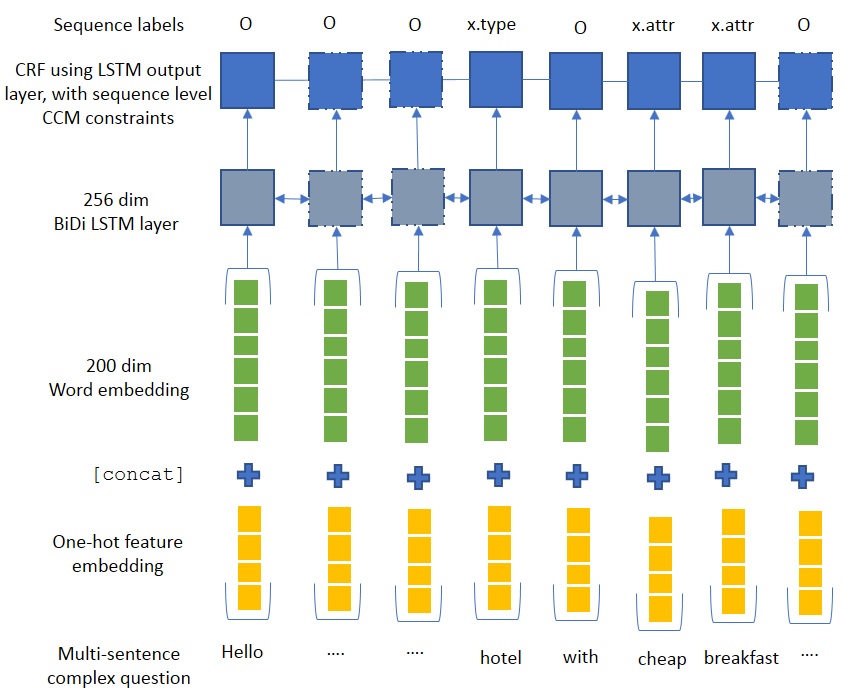}
  \caption{BiDi LSTM CCM for labeling question tokens.}\label{fig:bidi-LSTM}
}
\vspace*{-3ex}
\end{figure}

\noindent{\bf BiDi LSTM based sequence modeling:} We also experiment with neural approaches by modeling each question using a bi-directional LSTM CRF. 
The input states in the LSTM are modeled using a $200$ dimension word vector representation of the token. These word vector representations were pre-trained using the word2vec model\cite{word2vec} on a large collection of $80,000$ tourism questions. 

We extend the basic BiDiLST CRF model in two ways to improve performance in our low-data setting (see Figure \ref{fig:bidi-LSTM}). First, we allow use of hand-crafted features by representing each unique feature set as a one-hot vector and concatenating this feature vector with the word-vector representation of each token\footnote{We also experimented with using a feature vector in the CRF layer instead of the LSTM input layer but it gave poorer results}. Second, we enforce the CCM constraints during inference and the model is trained end-to-end using back-propagation. {\color{black} To the best of our knowledge, this combined model of BiDiLSTM CRF with CCM constraints is a novel contribution of our work.}

\subsubsection{Crowd-sourced Data Collection}
\label{sec:data}

In order to obtain a larger amount of labeled data for our task, we decided to make use of crowd-sourcing (Amazon Mechanical Turk). Since our labeling task can be fairly complex, we divide our crowd task into multiple steps. We first ask crowd to (i) filter out forum questions that are not recommendation questions. For the questions that remain, the crowd provides (ii) $user.*$ labels, and (iii) $x.*$ labels. For each step, instead of directly asking for token labels, we ask a series of indirect questions that lead us to the desired labels. For example, for $x.type$, we ask workers to highlight the text that describes what the user is asking for as shown below:

\begin{itemize}
{
\item {\it ``Which sequence of words in the question tells you what the user is asking for? Label only one sequence from a single sentence; we prefer a continuous sequence."}
\item {\it ``What is the shortest sequence of words in ``A1 (answer to the question above)'' that describes a category (e.g. place to stay, restaurant, show, place to eat, spot, hotel etc.)?''}
}
\end{itemize} 

\begin{table}[t]
\centering
{\small
\begin{tabular}{|l|ccc|}
\hline
                               & $x.type$  & $x.attr$ & $x.loc$ \\ \hline

Token level disagreement (\%) & 52.02 & 62.22   & 31.44    \\ \hline

\end{tabular}
\caption{Inter-worker disagreement on AMT}
\label{tab:annotator}
}
\end{table}
This alternate way of labeling by asking a series of questions 
is inspired by \cite{Luke}. We obtain two sets of labels (different workers) on each question. Unfortunately, despite breaking the task into simpler steps, we see disagreement among the workers on some labels (see Table \ref{tab:annotator} for token-level disagreement statistics). 
Some of the disagreement results from labeling errors due to complex nature of the task. In other cases, the disagreement results from their choosing one of the several possible correct answers. E.g., in the phrase ``{\it good restaurant for dinner}'' one worker labels $x.type=$`restaurant', $x.attribute=$`good' and $x.attribute=$`dinner', while another worker simply chooses the entire phrase as $x.type$. We find it difficult to relate expert-guidelines to the crowd in a succinct and clear manner.

Since most of our posts have some disagreement among the workers, how do we incorporate this data into our supervised learning setting? We devise the novel idea of only keeping the labels where the two annotators agree and disregarding the other labels, resulting in {\em partially labeled} sequences. .
This is different from disregarding the entire example where there is any disagreement, which would result in loss of significant (correctly labeled) data.
We use this partial supervision, along with our original training set, in a semi-supervised learning setting, to learn the parameters of our CCM model. To the best our knowledge, we are the first to exploit partial supervision from a crowd-sourcing platform in this manner.

\subsubsection{Semi-Supervised Labeling} \label{sec:semi-supervised}
In order to use these partially labeled sequences, we adapt the original CoDL algorithm to work with partial labels. We replace the unlabeled set $U$ by the partially labeled set; inference over the set involves predicting only the missing labels. This can be still be done using an ILP based formulation. Rest of the update equation remains the same. 
In the case of the BiDiLSTM CRF based formulation, the modeling remains the same as described, except the $\rm{Learn}$ function now corresponds to training the neural network via back-propagation.

{\footnotesize
\begin{table*}[thb]
\centering
\footnotesize
\begin{tabular}{lllll}
\hline
\multicolumn{1}{|l|}{\textbf{Model}}                          & \multicolumn{1}{l|}{\textbf{F1 (x.type)}} & \multicolumn{1}{l|}{\textbf{F1 (x.attr)}} & \multicolumn{1}{l|}{\textbf{F1 (x.loc) }} & \multicolumn{1}{l|}{\textbf{F1 (aggr)}} \\ \hline
\multicolumn{1}{|l|}{CRF (all features)}                        & \multicolumn{1}{c|}{51.4}          & \multicolumn{1}{c|}{45.3}          & \multicolumn{1}{c|}{55.7}         & \multicolumn{1}{c|}{50.8}      \\ 
\multicolumn{1}{|l|}{CCM} 
& \multicolumn{1}{c|}{59.6}          & \multicolumn{1}{c|}{50.0}          & 
\multicolumn{1}{c|}{56.1}         & \multicolumn{1}{c|}{55.2}      \\ \multicolumn{1}{|l|}{CCM (with all crowd data)}             & \multicolumn{1}{c|}{55.1}          & \multicolumn{1}{c|}{42.2}          & \multicolumn{1}{c|}{46.7}         & \multicolumn{1}{c|}{48.0}      \\
\multicolumn{1}{|l|} {Semi-supervised CCM}     & \multicolumn{1}{c|}{58.5}          & \multicolumn{1}{c|}{\bf 50.6}          & \multicolumn{1}{c|}{60.3}         & \multicolumn{1}{c|}{56.5}      \\ \hline
\hline
\multicolumn{1}{|l|}{BiDi LSTM CRF}             & \multicolumn{1}{c|}{53.3}          & \multicolumn{1}{c|}{47.6}          & \multicolumn{1}{c|}{52.1}         & \multicolumn{1}{c|}{51.0}      \\
\multicolumn{1}{|l|} {BiDi LSTM CRF with all features}     & \multicolumn{1}{c|}{58.4}          & \multicolumn{1}{c|}{48.1}          & \multicolumn{1}{c|}{62.0}         & \multicolumn{1}{c|}{56.2}      \\
 \multicolumn{1}{|l|} {BiDi LSTM CCM with Features }     & \multicolumn{1}{c|}{59.4}          & \multicolumn{1}{c|}{49.8}          & \multicolumn{1}{c|}{\bf 62.3}         & \multicolumn{1}{c|}{57.2}      \\
\multicolumn{1}{|l|} {Semi-Supervised BiDi LSTM CCM with features}     & \multicolumn{1}{c|}{\bf 62.9}          & \multicolumn{1}{c|}{ 50.4}          & \multicolumn{1}{c|}{ 61.5}         & \multicolumn{1}{c|}{\bf 58.3 }      \\
\hline

\end{tabular}
\caption{Sequence tagger F1 using  CRF with all features, CCM with all features \& constraints, and semi-supervised CCM over partially labeled crowd data. The second set of results mirror these numbers using a Bi-directional LSTM CRF. Results are statistically significant (paired t-test,p value$<$0.000124).
}
\label{tab:features}
\vspace*{-0.7cm}
\end{table*}
}

\subsection{Operator Post-processing \& RQL Generation} \label{sec:operators}
To construct the final RQL query, we need to identify the appropriate operator for each labeled phrase. For each operator, we start with a manually curated set of seed words, and expand it using synonym and antonym counter fitted word vectors \cite{counterfittedwv}. The resulting set of {\it trigger} words flag the presence of an operator in a sentence. A set of deterministic rules estimate the scope of each operator. Rules use semantic labels from sequence labeling, along with some token-level features such  as part of speech tags to identify scope. For instance, a token (or a set of continuous tokens with the same label) labeled by our sequence tagger that occur within a specified window of a trigger word for ``negation'', are in its scope.  A secondary set of rules are used to compose operators with each other. For instance, if a phrase that is in scope for ``negation'' is also in scope for a ``disjunction'', then the ``negation'' applies to all tokens in scope of the disjunction. Further, in this case the disjunction converts to a conjunction by laws of Boolean algebra.

\subsection{Evaluation} \label{sec:experiments}

This evaluation answers the questions: (1) which model obtains the best performance for our semantic tagger, and (2) what is the incremental contribution of features, CCM constraints and crowdsourced annotation for our task. {\color{black} We additionally learn about effectiveness of neural models for sequence labeling in low-data setting.}

\noindent{\bf Evaluation Details:} We use the 150 expert-annotated tourism forum questions as our dataset and perform leave-one out cross-validation. Our current implementation trains only a subset of the labels \{$x.type$, $x.attribute$, $x.location$, $other$\} -- these are most important for downstream QA.

We use the {Mallet toolkit\footnote{http://mallet.cs.umass.edu/}} for CRF implementation and the GLPK ILP-based solver\footnote{https://www.gnu.org/software/glpk/} for CCM inference. 

For expts with semi-supervised learning, we add 400 partially-annotated questions from crowdsourced workers to our training set. 
We retain token labels marked the same by two workers for every question, and treat the other labels as unknown. We pay a total of \$1.02 per question for this annotation. We still compute a leave one out cross-validation on our original 150 expert-annotated questions (complete crowd data is included in each training fold). 
For CoDL learning we set $\gamma$ to 0.9 as per original authors' recommendations.

Sequence-tagged tokens identify {\em phrases} for each semantic label, which become units for constructing the final RQL query. So, instead of reporting metrics at the token level, we compute a more meaningful joint metric over tagged phrases. We define a matching-based metric that first matches each extracted segment with the closest one in the gold set, and then computes segment level precision using constituent tokens. Analogously, recall is computed by matching each segment in gold set with the best one in extracted set. As an example, for Figure \ref{fig:queryExample}, if the system extracts ``convenient to the majority'' and ``local budget'' for $x.attribute$ then our matching-metric will compute precision as 0.75 (1.0 for ``convenient to the majority'' and 0.5 for ``local budget)''  and recall as 0.45 (1.0 for ``budget'', 0.0 for ``best'' and 0.364 for ``convenient to the majority ... like to see'').

\noindent{\bf Results:} Table \ref{tab:features} reports the performance of our semantic labeler under different configurations. We find that using a CRF based system using all features gives us an aggregate F1 of $50.8$.  The use of our CCM constraints 
have a significant impact on overall performance, raising aggregate F-score by over 4 points.  
Table \ref{tab:features} shows that the CoDL training procedure for semi-supervised CCM boost the aggregate F-scores by a further 1.5 points. In order to see how useful our semi-supervised approach is, we also train a CCM based model that makes use of the {\em complete} crowdsourced dataset for training, {\color{black}by adding conflicting labels for a question as two independent training data points.} 
As can be seen, without a semi-supervised approach, the noisy crowd-labeled sequences hurts the performance significantly.

We also repeated the experiments using a BiDi LSTM+CRF using only the pre-trained word-embeddings as input features. We were surprised to find that it performs at par with the vanilla CRF model that uses all our engineered features. This speaks to the effectiveness of neural models even in low data scenarios. Our further extension on adding our features to the neural model yield a 5 pt performance boost. Both CCM constraints and semi-supervision improve F-scores by one point each. Overall, our best model combines neural features, hand-designed features, CCM constraints with semi-supervised learning over partially annotated crowd data.

\noindent {\bf Effect of features:} We perform an ablation study over our feature sets to learn about incremental importance of each feature (detailed results omitted due to lack of space).  We find that descriptive phrases, and hand-constructed type and attribute indicators improve the performance of each label by 2-3 points. Word2vec features help type detection because $x.type$ labels often occur in similar contexts, leading to informative vectors for typical type words. Frequency of non stopword words in the post are an indicator of the word's relative importance, and the feature also helps improves overall performance.

{\footnotesize
\begin{table}[thb]
\centering
\footnotesize
\begin{tabular}{|l|c|c|c|}
\hline
\textbf{Algorithm}           & \multicolumn{1}{l|}{\textbf{Prec}} & \multicolumn{1}{l|}{\textbf{Recall}} & \multicolumn{1}{l|}{\textbf{F1}} 
\\ \hline
CRF (all features) & {66.9}                                    & 41.7	                                 & 51.4                               
\\ 
CCM (all features)   &   62.1                        & {57.2}                                 & {59.6}                         

\\ \hline \hline

BiDI LSTM CRF with Features & {54.1}                                    & 63.6	                                 & 58.4                               
\\ 
BiDi LSTM CCM with Features   &   55.1                        & {64.5}                                 & {59.4}                                  
\\ \hline
\hline

\end{tabular}
\caption{
(i) Precision and Recall of {\it x.type} with and without CCM inference.
}
\label{tab:PRAnalysis}

\end{table}
}

\noindent{\bf Effect of constraints:} A closer inspection of Table \ref{tab:features} reveals that the vanilla CRF configuration sees more benefit in using our CCM constraints as compared to the BiDiLSTM+CRF based setup (4pt vs 1pt). To understand why, we study the detailed precision-recall characteristics of individual labels; the results for $x.type$ are reported in Table \ref{tab:PRAnalysis}. We find that the BiDiLSTM+CRF based setup has significantly higher recall than its equivalent vanilla CRF counter-part while the opposite trend is observed for precision. As a result, since two of the three constraints employed by us in CCM are oriented towards improving recall\footnote{Recall that we require at least one {\em  x.type} (hard constraint) and prefer at least one {\em x.attr} (soft constraint)}, we find that they improve overall F1 more by finding tags that were otherwise of lower probability (i.e. improving recall). However, note that the semi-supervised CCMs have similar performance benefits in both configurations.

\noindent
{\bf Error Analysis: }
Identifying attributes is the toughest task among all semantic labels. Attributes may be associated with entities irrelevant in answering a query. E.g., in {\em "Staying in a fancy house, looking for a clean beach. Ideas??"}, `fancy' is associated with the `house', and is irrelevant to the beaches being queried for. Moreover, depending on the phrasing of a question, there is a confusion between attribute and type entities: `Moroccan Food' in {\em "Looking for Moroccan Food"} is a type; while in {\em "Looking for a restaurant with good Moroccan Food"}, we mark it is an attribute. This confusion is also reflected in Table \ref{tab:annotator}, where workers agree the least on attributes. 

\noindent{\bf Operator Labeling: } We also evaluate the operator labeling component from Section \ref{sec:operators}. We create rules for commonly occurring operators --  disjunctions, negations, and preferences.\footnote{Our current system ignores IN and SIMILAR operators}  The default operator between semantic labels is a conjunction.
Table \ref{tab:operators} reports the accuracy of our rules as evaluated by an author. The `Gold' columns denote the performance when using gold semantic label mentions. The 'System'
columns are the performance when using labels generated by our sequence tagger. We find a significant drop in detection of disjunctions and preferences. A detailed study reveals that nearly 70\% of all disjunction clauses occur in the context of attribute labels and almost all preferences are expressed for attributes. Since system's recall for $x.attr$ is low, the pipeline under-performance is not entirely surprising.\footnote{There is a more severe drop in disjunctions because for a valid disjunction match we need at least {\em two} sets of $attribute$ phrases correctly identified in the same query.}  {\color{black}However, as the next section shows, the low recall for these tags is not very problematic for the end QA task, as they are relatively infrequent. Moreover, even if we miss some attributes, the other attributes can often lead us to correct answers.}

\begin{table}[]
\centering
{\footnotesize
\begin{tabular}{l|ll|ll}

             & \multicolumn{2}{c|}{Gold} & \multicolumn{2}{c}{System} \\ 
             & P        & R               & P         & R              \\ \hline \hline
Negations    &  86        &    66            &  85         &      62            \\ 
Preference   &   75      &    60      &   33        &  15               \\ 
Disjunctions &    86     &    59              &     50     &       08      \\ 
\vspace*{-2ex}
\end{tabular} 
}
 \caption{Performance of operator detection using gold sequence labels, and system generated labels}
 \label{tab:operators}
 \vspace{-0.7cm}
\end{table}
\section{Answer Generation} 
\label{sec:end-task}
The answer generation component serves as a translation layer from an RQL parse to the schema and querying mechanism of the knowledge source for retrieving entity answers. 
To demonstrate the ease of integrating different types of knowledge sources,  we experiment with two different knowledge sources: 
(i) an Apache Lucene data store containing data from Google Places, reviews from Trip Advisor, and Booking.com as well as articles from WikiTravel,  resulting in data for {\color{black}480,000} unique entities, (ii) the full Google Places collection queried using its Web API. 

We randomly select 150 new unseen questions {\color{black}(different from the questions used in the previous section)}, from a tourism forum website and manually remove 45 of those that were not entity-seeking. For the remaining 105 questions our annotators manually look up the actual entity-answers for those questions and this forms our test set for the answering task. 
\noindent
{\bf Baselines:} 
As a baseline, we adapt and reimplement a recent tf-idf scheme (called WebQA) originally meant for finding appropriate Google results to answer questions posed in user forums \cite{complexwebQA2016}.
WebQA first shortlists a set of top 10 words in the question using tfidf computed over the set of all questions; it forms a query by selecting 3 words based on supervised training data. Since we don't have any supervised data, we select the top 3 words in the shortlist as the query (space separated). Further, we improve this to a WebQA-Manual system in which an expert chooses 3-4 best words manually.

\noindent
{\bf Lucene-based QA: } We construct the background knowledge source by indexing data from four sources. Using a list of 3500 cities that have a population over 100,000: (i) we query Google Places API\footnote{\em https://developers.google.com/places/web-service/} and obtain records on 405,000 entities (hotels, attractions, restaurants, and other service providers); each record contains the Google Places type,\footnote{\em https://developers.google.com/places/supported\_types} address, overall rating, map coordinates but only five user reviews. (ii) we collect hotel data from Booking.com and index upto 75 reviews per hotel, and (iii) we index reviews of attractions and places to visit for each city using TripAdvisor.com. This results in a total of 454,000 distinct entities. {\color{black} (iv) To answer questions where users are asking for a city or places around a city, we add the full collection of WikiTravel\footnote{\em http://www.wikitravel.org} containing 26,000 cities along with its text.  For each entity's textual data (reviews, WikiTravel articles etc), we remove stop words, lemmatize it, and index as a record into Apache Lucene. }

Data collected from Booking.com and TripAdvisor.com does not contain a default ``type'' field. We index such records by marking their Lucene types as ``lodging'' for hotels and ``point\_of\_interest'' for attractions, since these values are used by other records from Google Places to refer to such entities. All other entities from Google Places retain their Google place type as the Lucene type. {\color{black} We also associate each city entity with a special Lucene type called ``city''. We also store its geographical coordinates using the Google Maps API.}

Our answer generator needs to translate a question's RQL representation into a Lucene Query. It first selects the most appropriate Lucene type for the answer by calculating the cosine similarity between RQL $x.type$ phrase and all Lucene types in counter-fitted word vector embedding space \cite{counterfittedwv}.  It uses phrases from $x.attribute$ to query the text fields, and applies filters on location using $x.location$ and question meta-data. We also use the $x.type$ phrases to query the text fields, as sometimes descriptive phrases get included as part of the $x.type$ (e.g, ``budget hotel''). Negations, disjunctions and conjunctions are enforced appropriately using Lucene BooleanQuery\footnote{\em https://goo.gl/MYuH2L}. For now, we ignore PREF and other non-standard operators when forming the Lucene query. For fairness, the WebQA baselines also enforce location filters using question metadata and use in-built Lucene query parsers to generate queries. 

As a back-off strategy, in case the system returns no results (sometimes because of a very strict query), we relax the query by replacing conjuctions with disjunctions between $x.attr$ tags. For the baseline systems the back-off strategy drops least important (low tf-idf weighted) terms from the query.

\begin{table*}[]
\centering


{\footnotesize
\begin{tabular}{c|c|ccc}
{\bf Data} & {\bf System} & {\bf Acc@3} & %
{\bf MRR} & {\bf Recall}\\
\hline \hline

\multirow{3}{*}{\textbf{Lucene}}                                               & Default Lucene Query parser & 23.7 & 0.16 

& { 20.9} \\ 

& WebQA & 25  & 0.17 
& {21.9} \\ 

& WebQA (manual) & 27.7 & 0.23
& {23.8 } \\ 

& RQL-QA& { 34.6} & 0.28
& 24.8 \\

\hline
\multirow{2}{*}{\textbf{}}  & WebQA                                                           & 50 & 0.47 
& 4.7 \\
{\bf Google} & WebQA (manual) & 42.6 & 0.39 
& 30 \\ 
{\bf Places API} & RQL-QA & {\bf 62.5} & { \bf 0.52} 
& {\bf 47.6}

\end{tabular}
\vspace{-0.3ex}
\caption{QA results on two knowledge sources}
\label{tab:qa-task}
}
\vspace{-1cm}
\end{table*}

\begin{table*}[]
\footnotesize
\begin{tabular}{|p{0.2cm}|p{9cm}|p{1.5cm}|p{5cm}|}
\hline
{\bf No.} &{\bf Question}                                                                                                                                                                                                                                                                                                                                                                                                                                                      & {\bf Entity Type}  & {\bf System Answer}
\\ \hline \hline
1&My family and my brother's family will be in Salzburg over Christmas 2015. We have arranged to do the Sleigh Ride on Christmas day but are keen to do a local style Christmas Day dinner somewhere. Any suggestions?                                                                                                                                                                                                                                            & Special Dinner place                       &
{\color{cadmiumgreen}St. Peter Stiftskulinarium, {\tiny Sankt-Peter-Bezirk 1\/4, 5020 Salzburg}}
\\ \hline
2& Heading to Salzburg by car on Friday September 18th with my wife and her parents (70's) and trying to make the most of the one day. Thinking about a SOM tour, but not sure what the best tour is, not a big fan of huge groups or buses, but would sacrifice for my Mother in Law (LOL). Also thinking about Old Town or the Salzburg Fortress. Any suggestions? Where to park to have easy access as well as a great place for dinner.Thanks so much! & Tour                                       & 
{\color{cadmiumgreen} Bob's Special Tours, {\tiny Rudolfskai 38, 5020 Salzburg, Austria}}
\\ \hline
3& What can you do in Helsinki on a Sunday morning? What would you recommend a tourist to do or see on a Sunday morning? I'll be ariving at 7 in the morning, and it seems like everything's closed on a Sunday morning- either its not open on Sundays or else it'll open but later on in the day.                                                                                                                                                         & Things to do / see & 
{\color{cadmiumgreen} \shortstack{ Senate Square, {\tiny 00170 Helsinki, Finland}\\ Ateneum,{\tiny Kaivokatu 2, 00100 Helsinki, Finland} \\ ..}}

\\ \hline
4 &I am planning to visit Agra for 2 days in mid Dec with my friends.My plan is to try some street food and do some local shopping on day 1 and thus wish to stay in a good budget 3 star hotel (as I wont be spending much time in the hotel) at walking distance from such street food\/local shopping market.Then on the 2nd day, I want to just relax and enjoy the hotel.(I have booked a premium category hotel, Radisson Blu for this day hoping for a relaxed stay)Please suggest some good hotel or market around which I should book an hotel for my first day. & Hotel with location constraints & 
 {\color{red} Hotel Taj Plaza, Agra, {\tiny Taj Mahal East Gate, Near Hotel Oberoi Amar Vilas, VIP Road, Shilpgram, Agra, Uttar Pradesh 282001, India}}
 
 \\ \hline
5 & Hi there. I am going to Tallinn in a month from just one night on a Saturday. I am 28 and am going with 5 of my friends. Were should we stay so we are near the best clubs in the city? Any recomendations are appreciated!!! Thanks.& Place to stay close to clubs & 
 {\color{red} Club Prive, {\tiny Tallinn, Estonia}}
 \\ \hline
6&A few friends and I are coming up to Newport for a couple of nights and are looking for restaurant suggestions. We are thinking something casual for the first night. Is Flo's any good? And then something nicer on Saturday night....preferably a restaurant with good seafood. Also, any suggestions for good breakfast?                                                                                                                                     & Restaurant based on cuisine                & 

{\color{cadmiumgreen} The Red Parrot Restaurant, {\tiny  48 Thames St, Newport, RI 02840, United States }}
\\ \hline
7&Dear All forum members, I am Yash Khatri from Delhi.I am travelling to Srinagar on 13th July,2016 to 17th July,2016.I am going there for a show, and I'll be free on 15th and 16th July, 2016. I was thinking to hire a bike at Srinagar and travel toGulmarg\/Pahalgam.Queries :1) Where can I rent a bike at Srinagar and how much will it cost me?2) What is better for a quick visit; Gulmarg or Pahalgam?Please help!Thanks                          & Motorcycle rental                          & 
{\color{cadmiumgreen} Kashmir Bikers - Bike Rentals,  { \tiny Sheikh complex , shiraz chowk ,khanyar, Near j\&k bank khanyar, Srinagar, Jammu and Kashmir 190003}}
\\ \hline
8.& In a couple of weeks, we will have an almost 2 hr layover in Zagreb before flying on to Dubrovnik. Any recommendations for lunch ?& A location for lunch that can be visited in a 2 hour layover& 
     
{\color{red} Hotel Dubrovnik,{\tiny Gajeva ul. 1, 10000, Zagreb, Croatia}} 
     
     \\ \hline
9.&Hi,I am looking for a good hotel in Shillong (preferably near Police bazar) which would offer free wifi, spa and are okay with unmarried couples. My budget is 3k maximum. please suggest the best place to stay.                                                                                                                                                                                                                                               & Hotel with location and budget constraints &

 {\color{cadmiumgreen} {Hotel Pegasus Crown, {\tiny Ward's Lake Road, Police Bazar, Shillong, Meghalaya 793001, India ;
 }}
 } 
  \\ \hline
10.&Coming to Gent soon and we will take the train\/bus from Charleroi but ideally would like a taxi back from Gent to Charleroi. Can anyone recommend a good taxi firm please?                                                                                                                                                                                                                                                                                     & Taxi Service                               & 

{\color{cadmiumgreen} Taxi Didier Ghent Taxi Service, {\tiny Salviastraat 17, 9040 Sint-Amandsberg Gent, Belgium}}
\\ \hline
                                                                                                                                                                                                                                                                                                                                                                                                                                                               
\end{tabular}
\caption{Some sample questions from our test set and the answers returned by our system. 
Answers in {\color{cadmiumgreen} green} are identified as correct while those in {\color{red} red} are incorrect.}
\label{tab:questions}
\vspace*{-1cm}
\end{table*}

In order to answer queries that have $NEAR$ operators, a Lucene query is first generated to identify candidate {\color{black} locations (of type City)} by using the map co-ordinates and the Lucene Geo-point distance query API\footnote{\em https://goo.gl/mmrUcj}. A second query is then generated that selects the best entities {\color{black} from these cities} using the $x.attribute$ and the entities' text fields. This example demonstrates how some questions may require specialized query generation and our RQL representation enables easy translations into native KB query language.

\noindent
{\bf Google Places based QA: } {\color{black} While Lucene store has a large number of entities, it often has very few reviews for many of them. This may often miss the reviews necessary to match with attributes expressed in the question. For the second setting, our answer generator translates an RQL parse directly into a Google Places API query. This makes use of all of Google Places knowledge, as well as its advanced IR capability, while offering less control on the query langugage.}

We generate Google Places query via the transformation: ``concat ($x.attribute$) $x.type$ in $x.location$'', with operators applied at appropriate labels.  Here, concat lists all attributes in a space-separated fashion. This query under-exploits our rich semantic representation, but still returns more answers than Lucene due to its coverage. 

In case the API returns no results, we relax our query by dropping $x.attributes$ and then re-querying Google Places. This is sometimes helpful because RQL may contain a lot of $x.attribute$ tags that overwhelm the API\footnote{$x.attribute$ tags with values such as ``good', ``great'', ``best'', ``convenient'' are non-informative and the Google Places API internal ranking implictly ranks for adjectives even if they aren't specificed in the query. An explicit specification of too many such tags can cause the API to return no results.}. A downside of dropping $x.attribute$ tags, however, is that any strict selectional preferences expressed by the user in the question may be lost because we don't identify if some $x.attribute$ tags are more important than others.

\noindent{\bf Results:} Table \ref{tab:qa-task} reports Accuracy@3, which gives credit if any one of the top three answers is a correct answer. It also reports Mean Reciprocal Rank (MRR). Both of these measures are computed only on the subset of attempted questions (any answer returned). Recall is computed as the percentage of questions answered correctly within the top three answers over all questions. In case the user question requires more than one entity type\footnote{A question can ask for multiple things, eg., `museums' as well suggestions for ``hotels''.}, we mark an answer correct as long as one of them is attempted and answered correctly.  

For Lucene, while all systems answer nearly equal number of questions, RQL-QA has a 7-11 pt point higher accuracy. This is because of the type constraints enforced while querying. Some of its errors are due to incorrect matching of textual types (e.g., `things to do') to KB types (tour agencies) due to faulty word-vector distances.

RQL-QA for Google Places answers many more questions than Lucene-based QA  because the online API has high review coverage. Moreover, its query processor is likely more sophisticated than Lucene's,  for e.g., in handling types like `things to do'. RQL-QA has an 20 point higher accuracy with a 17 point higher recall compared to WebQA (manual), because of a more directed and effective query to Google Places API.  

\begin{table*}
\centering
\footnotesize
\begin{tabular}{|p{3cm}|p{1cm}|p{4.2cm}|}
\hline
{\bf Error Type}                                                 & {\bf Error (\%)} & {\bf Examples}                                                                                                                                                                                                           \\ \hline \hline
Incorrect answer returned due to incorrect $x.type$ in RQL & 16            & Bad $x.type$ extractions in RQL results in incorrect answers.                                                                                                                                                       \\ \hline
Incorrect answer returned by knowledge source & 21           & $x.attribute$ criteria was not fulfilled - eg. Shop allows renting bicycles but not for tours.                                                                                                                     \\ \hline
Incorrect answer returned due to incomplete RQL & 10            & $x.attribute$ not getting extracted                                                                                                                                                                                 \\ \hline
Answer not returned by knowledge source & 21  & No apparent errors in RQL and the knowledge base would have been expected to be able to answer such a query.                                                                                                        \\ \hline
Answer not returned due to RQL errors & 16  & RQL errors such as bad $x.attribute$, $x.location$, too many or incorrect $x.type$ etc.                                                                                                                             \\ \hline
Answer not returned due to knowledge source limitations & 16            & Query requesting places ``around'' a city, or between two cities, $x.type$ extracted as ``day trips'', ``cruises'' etc. Requests for $x.type$ where queries were about bus services, activities and train stations. \\ \hline
\end{tabular}
\caption{Classification of errors made by our answering system (using Google Places web API as knowledge source)}
\label{tab:errors}
\vspace{-0.7cm}
\end{table*}
Table \ref{tab:questions} shows some questions and the top-ranked answer entity returned by our system. 
As can be seen our system supports a variety of question intents/entities and due to our choice of an open semantic representation, we are not limited to specific entity types, entity instances, attributes or locations. For example, in $Q1$ the user is looking for ``local dinner suggestions'' on Christmas eve, and the answer entity returned by our system is to dine at the ``St. Peter Stiftskulinarium'' in Salzburg, while in $Q2$ the user is looking for recommendations for ``SOM tours'' (Sound of Music Tours). A quick internet search shows that our system's answer, 'Bob's Special Tours', is famous for their SOM tours in that area. This question also requests for restaurant suggestions in the old town, but since we focus on returning answers for just one $x.type$, this part of the question is not attempted by our system. Questions with more than one $x.type$ requests are fairly common and this sometimes results in confusion for our system especially if $x.attribute$ tags relate to different $x.type$ values. Since we do not attempt to disambiguate or link different $x.attribute$ tags to their corresponding $x.type$ values, this is often a source of error. Our constraint that forces all $x.type$ labels to come from one sentences mitigates this to some extent, but this is can still be a source of errors. 
$Q9$ is a complicated question with strict location, budget and attribute constraints and the top ranked returned entity ``Hotel Pegasus Crown'' fulfills the most requirements of the user\footnote{The hotel does not offer a spa and even with manual search we could not find a better answer}. 
$Q4$ is incorrect because the entity returned does not fulfil the location constraints of being close to the ``bazar'' while $Q5$ returns an incorrect entity type.

\noindent{\bf Error Analysis:} As can be seen in Table \ref{tab:qa-task} our best system attempts approximately 47\% of the questions with an acceptable degree of accuracy for the challenging task of answering MSRQs.  We conducted a detailed error study on our test set of 105 questions which is summarized in Table \ref{tab:errors}. We find that approximately 37\% of questions were not answered by our system due to limitations of Google Places, i.e. either an answer was not returned for unknown reasons or the question was un-answerable with the data available in the knowledge source. Another 21\% of the questions were answered incorrectly by the knowledge source, sometimes due to shallow query translation from RQL, while approximately 42\% of the recall loss in the system can be traced to errors in the RQL representation.

\vspace{-1ex}
\section{Conclusion and Future Work} \label{sec:conclusion}
We have presented the novel task of answering entity-seeking multi-sentence recommendation questions in the tourism domain. As a first solution, we proposed a pipelined model consisting of two steps: (a) Question Understanding (b) Question Answering. We proposed an SQL-like query representation for capturing the semantic content of a question. We formulated the task of generating the semantic representation as a sequence labeling task and presented a CRF based model using BiDiLSTM based as well as hand-tuned features, trained in a semi-supervised setting. Our model explicitly makes use of constraints. For answering, we have proposed to construct knowledge source specific queries from our question representation, which are fired over underlying knowledge sources. We have presented an end-to-end evaluation of our system over two answer repositories, showing that our model significantly outperforms the baseline models.

We see our paper as the first attempt towards end to end QA in the challenging setting of multi-sentence questions answered directly on the basis of information in large textual corpora. It opens up several future research directions, which can be broadly divided in two categories. First, we would like to improve on the existing system in the pipelined setting. Error analysis on our test set suggests the need for a deeper IR system that parses constructs from our semantic representation to execute multiple sub-queries. Currently, between 37-58\% of recall loss is due to limitations in the knowledge source and query formulation, while a sizeable 42\% may be addressed by improvements to question understanding. 

As a second direction, we would like to train an end to end neural system to solve our task. This would require generating a large dataset of labeled QA pairs which could perhaps be sourced semi-automatically using data available in tourism QA forums. However, answer posts in forums can often refer to multiple entities and automatically inferring the exact answer entity for the question can be challenging. Further, we would have to devise efficient techniques to deal with hundreds of thousands of potential class labels (entities). Comparing the performance of the pipelined model and the neural model, and examining if one works better than the other in specific settings would also be interesting to look at. 

Exploring other question types such as suggestions for itineraries, fact-check or yes-no questions, experimenting in different domains such as consumer electronics, automobiles etc could also be an interesting direction of future work. 

\section{Acknowledgements}
We would like to thank Poojan Mehta who designed and setup the annotation tasks on Amazon Mechanical Turk. We would also like to acknowledge the IBM Research India PhD program that enables the first author to pursue the PhD at IIT Delhi. This  work  is  supported  by  Google  language  understanding  and  knowledge  discovery  focused  research  grants,  a Bloomberg  award,  a  Microsoft  Azure  sponsorship, and a  Visvesvaraya  faculty  award  by  Govt. of India  to  Mausam. Parag is being supported by the DARPA Explainable Artificial Intelligence (XAI) Program under contract number N66001-17-2-4032. We thank all AMT workers who participated in our tasks.

\bibliography{acl2017}
\bibliographystyle{ACM-Reference-Format}
\end{document}